\pdfoutput=1

\documentclass[11pt]{article}

\usepackage{acl}

\usepackage{times}
\usepackage{latexsym}

\usepackage[T1]{fontenc}

\usepackage[utf8]{inputenc}

\usepackage{microtype}

\usepackage{tipa}
\usepackage{dingbat}
\usepackage{linguex}

\title{Norm Participation Grounds Language}

\author{David Schlangen \\ %
  CoLabPotsdam / Computational Linguistics \\
  Department of Linguistics, University of Potsdam, Germany\\
\texttt{david.schlangen@uni-potsdam.de}
}

\begin{document}
\maketitle

\begin{abstract}
  The striking recent advances in eliciting seemingly meaningful language behaviour from language-only machine learning models have
  only made more apparent, through the surfacing of clear limitations, the need to go beyond the language-only mode and to \emph{ground} these models ``in the world''. Proposals for doing so vary in the details, but what unites them is that the solution is sought in the addition of %
  non-linguistic data types such as images or video streams, while largely keeping the mode of learning %
  constant.
  I propose a different, and more wide-ranging conception of how grounding should be understood: What grounds language %
  is its normative nature. %
  There are standards for doing things right, these standards are public and authoritative, while at the same time acceptance of authority can and must be disputed and negotiated.
  What grounds language, then, is the determined %
use that language users make of it, and what it is grounded in is the community of language users. I sketch this idea, and draw some conclusions for work on computational modelling  of meaningful language use.
\end{abstract}

\section{Introduction}

You fire up the lastest language model---a machine learning model induced by guessing words in a very large body of text---and ``ask'' it: ``How would you describe the colour green?''. After a short pause, the following text appears: ``\emph{The colour green is refreshing and vibrant. It is often associated with nature and growth.}''. You: ``How would you describe the colour red?'' -- ``\emph{The colour red is exciting and full of energy. It is often associated with passion and power.}''. Not bad. You hold two objects up in front of the computer screen: ``Which of these is green?''. The reply suddenly is less than satisfying (and also old news in this conversation): ``\emph{The colour green is refreshing and vibrant. It is often associated with nature and growth.}''\footnote{%
  Output by the GPT-3 model (\texttt{text-davinci-002}) by openAI \citep{gpt3}, retrieved on 2022-05-18. %
}

This of course is a profoundly unfair test. The model has no connection to you other than through what you type, and so can't observe what ``these'' refers to, and moreover, it \emph{never} had access to anything other than language data.
The question then arises what the consequences of this limitation are. Is it just that those models (unsurprisingly) can't tell one visually presented object from another, or is there something fundamentally off about their grasp of language?

In an influential recent paper, \citet{Bender2020} argue that indeed something is off, in that a model that only has access to form cannot learn to connect that form to meaning, which they tie to communicative intention, which in turn can be \emph{about} the world. They also suggest that perhaps giving access to more than linguistic form might make it possible to learn this connection. 
In this their argument meets with more mainstream views that don't deny meaning status to language-only (or ``internet world scope'') models, but see them as deficient until augmented with additional forms of data \cite{Bisk2020}.

Here, I will argue that just connecting language with non-language data still misses fundamental properties of language use, along two dimensions. First, the connection between world states and language is only one among several types of connections that must be gotten right (the others being intra-language connections, and connections between language and actions). Secondly, the connections themselves need to be understood as \emph{normative} ones, which again has two kinds of consequences: They effect \emph{commitments} and \emph{entitlements}, but they also are non-necessary and their applicability must be argued for (and can be argued against).
Just getting things right occasionally, or even very often, is not enough. The getting it right must come from an orientation towards the relevant standards of getting things right. This ``orientation towards'' shows in the ability to appeal to these standards when challenged, either directly or in the repair of understanding problems, and it forms the difference between what I will call ``\emph{norm conformance}'' (which is what AI models are trained for), and ``\emph{norm participation}'' (which is what what grounds language use). %
In short, there are interactive capabilities that need to underwrite meaningful language use, rendering agents that do not have them deficient language users. As I will argue, this has consequences for the use of NLP systems that can falsely appear as having these capabilities, and it also opens up interesting research directions.

Let us start by looking at simple observational statements, and use this to draw a schematic picture of meaning making in language use.

\section{``There is a tiger.''}
\label{sec:tiger}

``There is a tiger'', you say, facing your friend but looking past them at a location behind their back. Leaving aside what these news should do to your friend, let's think a bit about what I, an overhearer, am justified to think may have been done to you to make you say that, and how that allows me to assign meaning to what you said.

So, why did you say that? There are many possible types of answers to this question, a central class among which (namely those that not also insinuate malicious intent to deceive on your side) will mention in some form the \emph{state of affairs} of there being a tiger, and your stating this as a fact. %
That is, there is an assumed connection between your expression $e$ or more generally your action $a$ (of uttering $e$) and a state of affairs $c$.
You said ``there is a tiger'' in part because there is a tiger.

But let's say you were wrong, and it was just Tibby, my oversized tabby cat which can look, at least for a split second and when she is very hungry, like a (still very small though) tiger. We can't say anymore that you said ``there is a tiger'' because there was a tiger, as there was none. But we can fix the description by giving you a---potentially misguided---inner life: you said ``there is a tiger'', because you believed there to be a tiger, and you believed there to be a tiger, because you misperceived Tibby as one. The chain now goes from $c$, the state of affairs (which in the modified example does not hold, i.e.\ turns out not to be a fact), to $b$, the belief, to $a$, the action.

This chain can also be used to reconstruct understanding.\footnote{%
  Note that the following does not describe a process model. It may very well be that in actual interpretation, shortcuts can be applied that identify the verbal action as part of a larger action type. What matters for the rational reconstruction here is that the constructs described here (beliefs, intentions) are available in reasons you can give for your actions, after the fact.
}
How can I get from observing $a$ to forming my own belief about $c$? To reverse this chain, I need to see $a$ as representative of a \emph{type} of action $A$, and I need to know something about this type's connection to a \emph{type} of belief $C'$, and its connection to a \emph{type} of states of affairs $C$---and I need to assume that you expected your addressees to know this and to be able to use this knowledge to reason back to the best explanation.\footnote{
  The knowledge has to be about types, since $a$, the actual physical event, has happened only now and never before and will never happen again, I cannot previously have known anything about it, other than what I know or learn about the type of which it is a token.
}

Let us write out these connections in the form ``if $C$, then $\_\_\_$ $A$'', where the underscore shall work as a placeholder for a predicate describing the force of the connection, to be explored presently. If we take the step of seeing the forming of beliefs as an action as well, then this schema covers both parts of the chain described above: ``if you are looking at a tiger (and your eyes are open, and you are sighted, etc.), then you $\_\_\_$ form the belief that there is a tiger'', and ``if you hold the belief that there is a tiger and you want to inform me of it, then you $\_\_\_$ say (something to the effect of) `there is a tiger' ''.

Looked at from a different perspective, your saying ``there is a tiger'' has \emph{committed} you to believing that there is a tiger (insofar as that this is the best explanation for why you said that), and to having a good justification for that belief, where the best justification would be there indeed being a tiger (and you having the right kind of epistemic standing).
Having this belief further commits you to having other beliefs as well, such as ``there is a mammal'' or ``there is a four-legged animal'', ``there is a cat-like creature'', ``there is a living entity'', etc.; these are just consequences (material inferences, to be precise) that we can see as being contained in having this belief, or, in other words, as contributing to individuating this belief as the one that it is.

To collect what we have before we move on: This analysis assumes that there are connections between ways the world is and beliefs about it, between beliefs and other beliefs, and between beliefs (and other mental states, such as intentions) and actions. (We leave aside here whether in an ultimate analysis, these beliefs could not be explained away as dispositions to act in a proscribed way.)
These connections can figure both in explanations of why you do something (you do $A$, because you are in $C$) and in abductions about states (as you did $A$, the state most likely is $C$).
What is open is the exact nature of these connections, which is what we will turn to next.\footnote{
  What I've tried to convey in this short section is my take on some Sellarsian themes \citep{sellars:languagegames,sellars:thought,devries:sellars}, especially with the three main moves of language-entry, intra-language movement, and language-exit \citep{sellars:languagegames}, and a conceptual-role semantics for propositional attitudes \citep{Harman1987-HARNCR}. This will need to be expanded on in more detail elsewhere.
}

\section{Norm Conformance and Norm Participation}
\label{sec:conf}

Our task now is to further specify the ``if $C$, then $\_\_\_$ $A$'' schemas. We want to achieve that they can figure in reasoning about
why a speaker said what they said, and, equally importantly, can be offered by the speakers themselves as reasons for why they said what they said. As we will see, these are related, but separate aims: Things can happen for a reason in different ways.

To make the discussion more concrete, let us instantiate the $C$ and $A$ in this schema, as follows:

\ex. \label{ex:tiger}
``if \emph{presented with visual features of this kind \leftpointright [picture of tiger]}, then you $\_\_\_$ \emph{say `there is a tiger'}''

Could this conditional feature in reasoning about the behaviour of a human speaker? We would probably hesitate to allow such a description, wanting to qualify the antecedent with something like ``\emph{and you want to inform your interlocutor about what you see, using the English language}'', for otherwise there are many ways in which you could react to the stimulus. Also, there is still the question of how to fill the placeholder, and it seems that it should indeed be filled somehow, as a conditional of the form ``if $C$, then you say $E$'' seems too strong as description of human linguistic behaviour; even wanting to do something does not unconditionally lead to doing it.

Before we come to that, however, we can observe that something like \ref{ex:tiger}, without any qualification about wanting to inform, is not a bad description of what the training set and learning objective of an image captioning model (e.g., \citet{mitchell-etal-2012-midge,vinyals:show}) realises: To the extent that the model works (as measured by accuracy, or some other metric that measures agreement with a reference), it \emph{conforms} to the \emph{norm} described by \ref{ex:tiger}. To the extent that this type of description fails to characterise the human language use situation, these models remain ungrounded.

What \ref{ex:tiger} misses, however, is that these regularities, these norms, can feature in self-explanations, and exert a stronger force on language users, which, I propose, is better expressed by making it an element of the norm: one \emph{ought to} behave in this way, given that the conditions are met; and, in reverse, one is \emph{committed to} them being met, if one behaved in this way. This opens up two possible points of contention in the application of such a norm: First, do the conditions indeed hold, that is, can it be applied? Second, is it even a norm, the authority of which I should accept? (``Says who?'' as a possible reply.) These are issues that can be, and not rarely are, raised in interaction (not in the artificial situations created by the language use of function-type models such as the aforementioned caption models). To distinguish this kind of actively following norms from just picking up regularities, I will use the label \emph{norm participation} for it.

Before we turn to the ways that this participation process plays out in interactive language use, let us unpack this proposal a bit more. Filling the placeholder and bringing in the intermediate belief state turns \ref{ex:tiger} into the following:

\ex. \label{ex:tigernorm}
\a. \label{ex:tigerbelief}
``if \emph{presented with visual features of this kind \leftpointright [picture of tiger]}, then you ought to \emph{believe that there is a tiger}''
\b. \label{ex:tigerb2b}
``if \emph{you believe that there is a tiger}, you ought to \emph{believe that there is a four-legged animal, and that there is a mammal, and that there is a living thing, and \dots}''
\b. \label{ex:tigersay}
``if \emph{you believe that there is a tiger, and you want to inform your interlocutor about this, using English}, you ought to \emph{say `there is a tiger'}''

To anticipate the discussion in the next section, the idea behind stressing the normative force of the connection is to explain why there is a pressure to correct disagreements, even if communicative success may have already been reached. In a very real sense, if you don't seem to be following these norms, then to me it seems that there is something wrong with you, at least as a participant in my \emph{system} of norms; or, potentially, there is something wrong with my system.

Let us start with a simple disagreement. Imagine you had pronounced \emph{tiger} as in German (/\textipa{"ti:g@}/); I can recognise which norm you were aiming for, but can point out to you that the correct form contains /\textipa{"taIg\textrhookschwa}/, which one ought to use. Or let us assume that I think that what is present is a leopard rather than a tiger; this allows me to spot that there is a deficiency in your belief/belief norm (and your perceptual one), at least compared to how I have it, which I can address by saying something like ``they look similar, but have a different coat: tigers have stripes, while leopards have spots''. (As we will discuss presently, I cannot force you to take this on; I can just try to make my claim of authority plausible to you.) Note that this limits the possible misunderstandings--- ``this is not a tiger, it's a gazelle'' is already odd; ``this is not a tiger, it's a refrigerator'' is far too odd, as the belief revision it indicates is too extreme to be plausible.

Lastly, an analysis of this form could also go some ways towards explaining why word uses can be so contentious, even if communicative success is not at issue: Each use makes the implicit claim that this is how one ought to talk, and that it makes the right kinds of distinctions, a claim that addressees may want to disagree with. An analysis of slurs and linguistic interventions \cite{McConnell-Ginet2020-MCCWMM,Cappelen2019-CAPBL-2} along these lines might be possible, but is left for future work here.

Again, let us take stock before we move on. I have argued that the right way to connect antecedent and consequent in constructs like \ref{ex:tiger} is to make  direct appeal to their normative status: it is not just that if $C$ is the case, one normally or conventionally does $A$, rather one \emph{ought to} do this, and does something wrong or at least something inviting correction when one does not do it. Doing things of these type then commits one in certain ways, and makes one suffer the consequences if these ways turn out to be not warranted. The analysis further has brought out a  distinction between (mere) \emph{norm conformance}, which is acting in accordance with a set of norms (for example, as they were realised in a data set of labelled examples) and %
\emph{norm participation}, which involves treating the norms as possible reasons for acting, which can be offered, requested, and challenged. The interactive processes in which this is done and which justify the label ``participation'' will be our topic next.

\section{Norm Participation as Interactive Process and Achievement}
\label{sec:inter}

The idea of the approach sketched here is that the question of which norms hold and how they are to be applied is never fully settled, and can become the overt topic of a conversation. That is, the fact that the connection is via an appeal to what one ought to do has practical consequences, which I will briefly trace in three related domains: language acquisition, conversational grounding, and conceptual disputes. More specifically, it shows in what in the field of conversational analysis is called \emph{repair}, and is rightly assigned a central place in the study of conversation \cite{schegetal:prefself,hayashi2013conversational,Jefferson2017b}.\footnote{%
  A recent cross-linguistic study by \citet{Dingemanse2015} found repair attempts on average about once per 1.4 minutes; studies of task-oriented dialogue found between 4 and 5.8\% of turns in the respective corpora to contain clarification requests \citep{purver:means,rodrischlang:catalog}.
  }

\paragraph{First Language Acquisition}

Children start out without knowledge of the norms of the language community in which they were born. Hence, they need to rely on the competent speakers around them to initiate them into these norms. The way they do this is by making attempts and observing reactions, which quite frequently involve \emph{repair}. For example, \citet{Golinkoff1986} found that about 50\% of attempts by small infants (in their first verbal phase, from 1 to 1.5 years old) resulted in repair. In the light of the schema proposed here, we can understand this as attempts at using a norm, being recognised as such, and then getting demonstrated how the act ought to be performed. As the examples collected by \citet{Clark2020} show, this process can target both the form (that is, schemata of the type of \ref{ex:tigersay}) as well as conceptual ones (as in \ref{ex:tigerbelief} and \ref{ex:tigerb2b}); indeed, these levels might often be addressed simultaneously. We can take away from this short review that an orientation towards shared norms seems to play a role already in the acquisition of language abilities.

\paragraph{Conversational Grounding}

According to H.~\citeauthor{clark:ul}'s (\citeyear{clark:ul}) well-known proposal, it is a constant task in conversation to ensure that common ground is reached, sufficiently for the purposes at hand. We can recognise the stages of this process (presenting \& identifying behaviour and signal; signalling \& recognising propositional state; proposing \& considering joint project) in the schema in \ref{ex:tigernorm}. More directly related are the consequences of successful conversational grounding, as discussed by \citet{brenclark:conpact} under the label ``conceptual pacts''. In the analysis proposed here, these can be understood as ``local'' norms that are not yet generalised, that is, ways the participants in an interaction mutually have come to think they ought to act with each other.\footnote{%
  A computational model of how such local conventions can reach whole populations has recently been offered by \citet{DBLP:journals/corr/abs-2104-05857}.
  }

\paragraph{Meaning Disputes}

The status of these norms as reasons for acting shows most clearly in those rarer cases where they need to be overtly discussed. Very occasionally, this can even become positive law: In \citep{tomato:vegetable}, the US Supreme Court judged that for the purposes of taxation, tomatos are vegetables, despite biologically better fitting under the label fruit. In our framework, this can be understood as an adjudication between a norm that better fits to one type of belief/belief system (tomatos as fruit, for biological reasons) vs.\ one that better accords to actual usage (tomatos as vegetables, for similarity in properties to other vegetables). Further examples are discussed by \citet{ludlow:living}, and more recently, under the label \emph{word meaning negotiation}, by \citet{larssonmyren:semco} and \citet{myrendal19:discstud}, who also provide the beginnings of a formalisation of the dialogue moves that structure this process.

\vspace*{\baselineskip}
\noindent
We can take from this very brief review that what is called norm participation here makes up a substantial amount of overt conversational moves, and is something that participants in verbal interactions actively engage in.

\section{Some Conclusions for Computational Modelling of Language Use}
\label{sec:lu}

I contrasted above \emph{norm conformance} from \emph{norm participation}, claiming that current natural language processing systems are only capable of the former, being optimised for \emph{accuracy} and not for systematic engagement in the processes reviewed in the previous section. A possible objection now is to reject that there is a problem---if accuracy can be raised sufficiently high, there would be no need for repair, and norm conformance would be indistinguishable from norm participation. This however presupposes that there is only one set of correct norms, and that this can in principle be found in the source datasets against which accuracy is measured. This is, however, is unlikely to be the case, once one moves outside of the very few domains with authoritative taxonomies (like an outsider may imagine Biology to work; \citet{Dupre2021-DUPTMO-6})---imagine a category like ``weed / pest plant''. The ``myth of the gold label'' is increasingly being noticed as a problem in NLP as well \citep{basile:perspectivist,Pavlick2019}.

If the story sketched above is on the right track, it provides a way to understand some ethical issues in the use of NLP systems.\footnote{
  These will be expanded in a separate paper, which will need to more thoroughly connect to the ongoing discussion in the nascent field of ``responsible AI''.
}
  Consumers of computer speech acts will assume that, just like with human speakers, something like the chain in \ref{ex:tigernorm} is in place in a captioning system for example, even if in reality there is a more direct and simpler link between visual input and language. A disagreement with a labelling decision or apparent category will need to find an addressee, which the system cannot provide. Organisations deploying such systems will need to take the responsibility for the ``commitments'' made by the system, as the system cannot do so -- as it cannot ``suffer the consequences''. Secondly, in the framework sketched above, as discussed, every use of language implicitly contains the claim ``this is how one does this''; again, on the principle that the system provider will need to pick up ``commitments'' made by the system, this is something that seems to argue against the deployment of language generation systems that are wont to reproduce undesirable material \citep{benderetal:parrot}.

As a final example along these lines, consider the application of question answering. In the discussion above, I briefly mentioned the condition of needing to possess the right kind of epistemic standing to form beliefs (discussed in more detail by \citet{Goldberg:assertion}). This epistemic standing can be ``inherited'' in knowledge through testimony \citep{Gelfert:testimony}. Current search engines indirectly honour these mechanisms, by framing their job only as surfacing source material that provides its own reputational claims towards such epistemic standing. Recent attempts at treating large language models as knowledge bases for question answering (surveyed by \citet{llmaskb-review}), however, break these links without providing others, which renders the status of their replies problematic (a similar point is made by \citet{Shah2022} and \citet{Potthast2020}).

With these caveats in mind, some potentially productive lines of work can also be motivated from within the framework explored here. A language generating system that is able to maintain a coherent system of norms as described here, can use them to offer self-explanations, and can react to corrections, would go some way towards more grounded, and hence more meaningful, language use. Components of this are already being explored separately. \citet{zhou-etal-2022-think} show that it is possible to explicate implicit commonsense knowledge from large language models (corresponding to the middle step  \ref{ex:tigerb2b}); \citet{kassner-etal-2021-beliefbank} show that a neuro-symbolic system can keep track of corrections to ``beliefs'' extracted from such models. It seems that combining these approaches in an interactive fashion, adding moves such as discussed by \citet{larssonmyren:semco}, would at least go some ways towards systems with more understandable meaning norms.

\section{Related Work}
\label{sec:relwo}

The inspiration from the work of Sellars for the ideas explored here has already been mentioned. Beyond the work cited above, the role that \emph{giving and asking for reasons} plays has been noted by \citet{sellars:epm} and explanded upon by \citet{brandom:explicit}.\footnote{%
  ``[I]n characterizing an episode or a state as that of knowing, we are not giving an empirical description of that episode or state; we are placing it in the logical space of reasons, of justifying and being able to justify what one says'' \citep[§36]{sellars:epm}
}
The varieties of rule following of course are an important topos from \citet{Witt:PU-corr} (see \citet{Baker2009-BAKWRG-6,Kripke1982-KRIWOR}), as is the necessarily public nature of judgements on the applicability of norms (on this see also \citet{Hegel:phaen}). 
The notion of ``orienting towards'' is central in the field of Conversation Analysis.\footnote{%
  ``CA’s guiding principle is that interaction exhibits ‘order at all points’ [\dots] This orderliness is normative—--it is produced and maintained by the participants themselves in their orientations to social rules or expectations'' \citep[p.2]{hoey:ca}

}

On the computational side, \citet{schlangen:justification} makes some related points, although not yet under the normative framework explored here. \citet{destone:societal} made a similar point, and much more carefully (but also more restricted in scope). The forming of conceptual pacts is investigated with modern computational means by \citet{takmaz-etal-2022-less}. Work that could be enlisted for going towards norm participating has already been cited in the previous section.

\section{Conclusions}
\label{sec:conc}

In this paper I have sketched a view of language as the purposeful use of norms for acting (where acting includes the forming of beliefs), where these norms can serve as reasons, can be negotiated, challenged, modified, and locally formed. I have speculated about the consequences of such a view on computational modelling of language use.

No one could mistake this offering here for more than a sketch. To develop this into a fuller proposal, an enormous amount of work remains to be done. How exactly language lends itself to figure in such norms, and how these are composed (note that all examples used full sentences) is an open question (and compositionality is notoriously a problem for conceptual role semantics \citep{whiting:crs}), to mention just one technical challenge.

Nevertheless, what I hope to have offered is a potentially productive way to think about how language is grounded, not just in some link to perceptual information, but in the collective uses made of it, which are actively constructed and maintained to be collectively useful. It is my hope that this more interactive perspective on symbol grounding can be informative for computational work on simulating language use.

\paragraph{Acknowledgements}

Many thanks to the anonymous reviewers for their very detailed and helpful comments. I would have liked to address them in more detail, but for reasons of time and space will need to do so elsewhere and some other time. This work was partially funded by the Deutsche Forschungsgemeinschaft (DFG, German Research Foundation) – 423217434 (RECOLAGE).

\bibliography{anthology,/Users/das/work/projects/MyDocuments/BibTeX/all-lit.bib}

\begin{thebibliography}{48}
\expandafter\ifx\csname natexlab\endcsname\relax\def\natexlab#1{#1}\fi

\bibitem[{AlKhamissi et~al.(2022)AlKhamissi, Li, Celikyilmaz, Diab, and
  Ghazvininejad}]{llmaskb-review}
Badr AlKhamissi, Millicent Li, Asli Celikyilmaz, Mona Diab, and Marjan
  Ghazvininejad. 2022.
\newblock \href {https://doi.org/10.48550/ARXIV.2204.06031} {A review on
  language models as knowledge bases}.

\bibitem[{Baker and Hacker(2009)}]{Baker2009-BAKWRG-6}
G.~P. Baker and P.~M.~S. Hacker. 2009.
\newblock \emph{Wittgenstein: Rules, Grammar and Necessity: Volume 2 of an
  Analytical Commentary on the Philosophical Investigations, Essays and
  Exegesis 185-242}.
\newblock Wiley-Blackwell.

\bibitem[{Basile et~al.(2021)Basile, Cabitza, Campagner, and
  Fell}]{basile:perspectivist}
Valerio Basile, Federico Cabitza, Andrea Campagner, and Michael Fell. 2021.
\newblock \href {http://arxiv.org/abs/2109.04270} {Toward a perspectivist turn
  in ground truthing for predictive computing}.
\newblock \emph{CoRR}, abs/2109.04270.

\bibitem[{Bender et~al.(2021)Bender, Gebru, McMillan-Major, and
  Shmitchell}]{benderetal:parrot}
Emily~M. Bender, Timnit Gebru, Angelina McMillan-Major, and Shmargaret
  Shmitchell. 2021.
\newblock \href {https://doi.org/10.1145/3442188.3445922} {On the dangers of
  stochastic parrots: Can language models be too big?}
\newblock In \emph{Proceedings of the 2021 ACM Conference on Fairness,
  Accountability, and Transparency}, FAccT '21, page 610–623, New York, NY,
  USA. Association for Computing Machinery.

\bibitem[{Bender and Koller(2020)}]{Bender2020}
Emily~M Bender and Alexander Koller. 2020.
\newblock {Climbing towards NLU: On Meaning, Form, and Understanding in the Age
  of Data}.
\newblock In \emph{Proceedings of the 58th Annual Meeting of the Association
  for Computational Linguistics (ACL)}, 2, pages 5185--5198.

\bibitem[{Bisk et~al.(2020)Bisk, Holtzman, Thomason, Andreas, Bengio, Chai,
  Lapata, Lazaridou, May, Nisnevich, Pinto, and Turian}]{Bisk2020}
Yonatan Bisk, Ari Holtzman, Jesse Thomason, Jacob Andreas, Yoshua Bengio, Joyce
  Chai, Mirella Lapata, Angeliki Lazaridou, Jonathan May, Aleksandr Nisnevich,
  Nicolas Pinto, and Joseph Turian. 2020.
\newblock \href {https://doi.org/10.18653/v1/2020.emnlp-main.703} {{Experience
  grounds language}}.
\newblock \emph{EMNLP 2020 - 2020 Conference on Empirical Methods in Natural
  Language Processing, Proceedings of the Conference}, pages 8718--8735.

\bibitem[{Brandom(1998)}]{brandom:explicit}
Robert Brandom. 1998.
\newblock \emph{Making it Explicit: Reasoning, Representing, and Discursive
  Commitment}.
\newblock Harvard University Press, Harvard, MA, USA.

\bibitem[{Brennan and Clark(1996)}]{brenclark:conpact}
Susan~E. Brennan and Herbert~H. Clark. 1996.
\newblock Conceptual pacts and lexical choice in conversation.
\newblock \emph{Journal of Experimental Psychology: Learning, Memory, and
  Cognition}, 22(6):1482--1493.

\bibitem[{Brown et~al.(2020)Brown, Mann, Ryder, Subbiah, Kaplan, Dhariwal,
  Neelakantan, Shyam, Sastry, Askell, Agarwal, Herbert{-}Voss, Krueger,
  Henighan, Child, Ramesh, Ziegler, Wu, Winter, Hesse, Chen, Sigler, Litwin,
  Gray, Chess, Clark, Berner, McCandlish, Radford, Sutskever, and
  Amodei}]{gpt3}
Tom~B. Brown, Benjamin Mann, Nick Ryder, Melanie Subbiah, Jared Kaplan,
  Prafulla Dhariwal, Arvind Neelakantan, Pranav Shyam, Girish Sastry, Amanda
  Askell, Sandhini Agarwal, Ariel Herbert{-}Voss, Gretchen Krueger, Tom
  Henighan, Rewon Child, Aditya Ramesh, Daniel~M. Ziegler, Jeffrey Wu, Clemens
  Winter, Christopher Hesse, Mark Chen, Eric Sigler, Mateusz Litwin, Scott
  Gray, Benjamin Chess, Jack Clark, Christopher Berner, Sam McCandlish, Alec
  Radford, Ilya Sutskever, and Dario Amodei. 2020.
\newblock \href {http://arxiv.org/abs/2005.14165} {Language models are few-shot
  learners}.
\newblock \emph{CoRR}, abs/2005.14165.

\bibitem[{Cappelen and Dever(2019)}]{Cappelen2019-CAPBL-2}
Herman Cappelen and Josh Dever. 2019.
\newblock \emph{Bad Language}.
\newblock Oxford University Press.

\bibitem[{Clark(2020)}]{Clark2020}
Eve~V Clark. 2020.
\newblock \href {https://doi.org/10.1080/0163853X.2020.1719795}
  {{Conversational Repair and the Acquisition of Language}}.
\newblock \emph{Discourse Processes}, 57(5-6):441--459.

\bibitem[{Clark(1996)}]{clark:ul}
Herbert~H. Clark. 1996.
\newblock \emph{Using Language}.
\newblock Cambridge University Press, Cambridge.

\bibitem[{DeVault et~al.(2006)DeVault, Oved, and Stone}]{destone:societal}
David DeVault, Iris Oved, and Matthew Stone. 2006.
\newblock Societal grounding is essential to meaningful language use.
\newblock In \emph{Proceedings of the Twenty-First National Conference on
  Artificial Intelligence (AAAI-06)}, Boston, MA, USA.

\bibitem[{DeVries(2005)}]{devries:sellars}
Willem~A. DeVries. 2005.
\newblock \emph{Wilfrid Sellars}.
\newblock Mcgill-Queen's University Press.

\bibitem[{Dingemanse et~al.(2015)Dingemanse, Roberts, Baranova, Blythe, Drew,
  Floyd, Gisladottir, Kendrick, Levinson, Manrique, Rossi, and
  Enfield}]{Dingemanse2015}
Mark Dingemanse, Se{\'{a}}n~G. Roberts, Julija Baranova, Joe Blythe, Paul Drew,
  Simeon Floyd, Rosa~S. Gisladottir, Kobin~H. Kendrick, Stephen~C. Levinson,
  Elizabeth Manrique, Giovanni Rossi, and N.~J. Enfield. 2015.
\newblock \href {https://doi.org/10.1371/journal.pone.0136100} {{Universal
  Principles in the Repair of Communication Problems}}.
\newblock \emph{Plos One}, 10(9):e0136100.

\bibitem[{Dupr\'e(2021)}]{Dupre2021-DUPTMO-6}
John Dupr\'e. 2021.
\newblock \emph{The Metaphysics of Biology}.
\newblock Cambridge University Press.

\bibitem[{Gelfert(2014)}]{Gelfert:testimony}
Axel Gelfert. 2014.
\newblock \emph{A Critical Introduction to Testimony}.
\newblock Bloomsbury Academic.

\bibitem[{Goldberg(2015)}]{Goldberg:assertion}
Sanford Goldberg. 2015.
\newblock \emph{Assertion: On the Philosophical Significance of Assertoric
  Speech}.
\newblock Oxford University Press.

\bibitem[{Golinkoff(1986)}]{Golinkoff1986}
Roberta~Michnick Golinkoff. 1986.
\newblock \href {https://doi.org/10.1017/S0305000900006826} {{‘I beg your
  pardon?': The preverbal negotiation of failed messages}}.
\newblock \emph{Journal of Child Language}, 13(3):455--476.

\bibitem[{Harman(1987)}]{Harman1987-HARNCR}
Gilbert Harman. 1987.
\newblock (nonsolipsistic) conceptual role semantics.
\newblock In Ernest LePore, editor, \emph{New Directions in Semantics}, pages
  55--81. London: Academic Press.

\bibitem[{Hawkins et~al.(2021)Hawkins, Franke, Frank, Smith, Griffiths, and
  Goodman}]{DBLP:journals/corr/abs-2104-05857}
Robert X.~D. Hawkins, Michael Franke, Michael~C. Frank, Kenny Smith, Thomas~L.
  Griffiths, and Noah~D. Goodman. 2021.
\newblock \href {http://arxiv.org/abs/2104.05857} {From partners to
  populations: {A} hierarchical bayesian account of coordination and
  convention}.
\newblock \emph{CoRR}, abs/2104.05857.

\bibitem[{Hayashi et~al.(2013)Hayashi, Raymond, and
  Sidnell}]{hayashi2013conversational}
Makoto Hayashi, Geoffrey Raymond, and Jack Sidnell. 2013.
\newblock \emph{Conversational Repair and Human Understanding}.
\newblock OAPEN Library. Cambridge University Press.

\bibitem[{Hegel(1807)}]{Hegel:phaen}
Georg Wilhelm~Friedrich Hegel. 1807.
\newblock \emph{Ph{\"a}nomenologie des Geistes}.
\newblock Philosophische Bibliothek. Meiner, Hamburg.
\newblock This Edition 1952.

\bibitem[{Hoey and Kendrick(2017)}]{hoey:ca}
Elliott~M. Hoey and Kobin~H. Kendrick. 2017.
\newblock \href {https://books.google.de/books?id=LXEzDwAAQBAJ} {Conversation
  analysis}.
\newblock In \emph{Research Methods in Psycholinguistics and the Neurobiology
  of Language: A Practical Guide}, Guides to Research Methods in Language and
  Linguistics. Wiley.

\bibitem[{Jefferson(2018)}]{Jefferson2017b}
Gail Jefferson. 2018.
\newblock \href
  {https://global.oup.com/academic/product/repairing-the-broken-surface-of-talk-9780190697969?q=Gail_Jefferson&lang=en&cc=nl}
  {\emph{Repairing the Broken Surface of Talk: Managing Problems in Speaking,
  Hearing, and Understanding in Conversation}}.
\newblock Foundations of Human Interaction. Oxford University Press, Oxford.

\bibitem[{Kassner et~al.(2021)Kassner, Tafjord, Sch{\"u}tze, and
  Clark}]{kassner-etal-2021-beliefbank}
Nora Kassner, Oyvind Tafjord, Hinrich Sch{\"u}tze, and Peter Clark. 2021.
\newblock \href {https://doi.org/10.18653/v1/2021.emnlp-main.697}
  {{B}elief{B}ank: Adding memory to a pre-trained language model for a
  systematic notion of belief}.
\newblock In \emph{Proceedings of the 2021 Conference on Empirical Methods in
  Natural Language Processing}, pages 8849--8861, Online and Punta Cana,
  Dominican Republic. Association for Computational Linguistics.

\bibitem[{Kripke(1982)}]{Kripke1982-KRIWOR}
Saul Kripke. 1982.
\newblock \emph{Wittgenstein on Rules and Private Language: An Elementary
  Exposition}.
\newblock Harvard University Press.

\bibitem[{Larsson and Myrendal(2017)}]{larssonmyren:semco}
Staffan Larsson and Jenny Myrendal. 2017.
\newblock {Dialogue Acts and Updates for Semantic Coordination}.
\newblock In \emph{semdial 2017}, pages 59--66, Saarbr{{\"{u}}}cken, Germany.

\bibitem[{Ludlow(2014)}]{ludlow:living}
Peter Ludlow. 2014.
\newblock \emph{Living Words}.
\newblock Oxford University Press, Oxford, UK.

\bibitem[{McConnell{-}Ginet(2020)}]{McConnell-Ginet2020-MCCWMM}
Sally McConnell{-}Ginet. 2020.
\newblock \emph{Words Matter: Meaning and Power}.
\newblock Cambridge University Press.

\bibitem[{Mitchell et~al.(2012)Mitchell, Dodge, Goyal, Yamaguchi, Stratos, Han,
  Mensch, Berg, Berg, and Daum{\'e}~III}]{mitchell-etal-2012-midge}
Margaret Mitchell, Jesse Dodge, Amit Goyal, Kota Yamaguchi, Karl Stratos,
  Xufeng Han, Alyssa Mensch, Alex Berg, Tamara Berg, and Hal Daum{\'e}~III.
  2012.
\newblock \href {https://aclanthology.org/E12-1076} {{M}idge: Generating image
  descriptions from computer vision detections}.
\newblock In \emph{Proceedings of the 13th Conference of the {E}uropean Chapter
  of the Association for Computational Linguistics}, pages 747--756, Avignon,
  France. Association for Computational Linguistics.

\bibitem[{Myrendal(2019)}]{myrendal19:discstud}
Jenny Myrendal. 2019.
\newblock \href {https://doi.org/10.1177/1461445619829234} {{Negotiating
  meanings online: Disagreements about word meaning in discussion forum
  communication}}.
\newblock \emph{Discourse Studies}, pages 1--23.

\bibitem[{{Nix v. Hedden, 149 U.S. 304}(1893)}]{tomato:vegetable}
{Nix v. Hedden, 149 U.S. 304}. 1893.

\bibitem[{Pavlick and Kwiatkowski(2019)}]{Pavlick2019}
Ellie Pavlick and Tom Kwiatkowski. 2019.
\newblock {Inherent Disagreements in Human Textual Inferences}.
\newblock \emph{Transactions of the Association for Computational Linguistics},
  7:677--694.

\bibitem[{Potthast et~al.(2020)Potthast, Hagen, and Stein}]{Potthast2020}
Martin Potthast, Matthias Hagen, and Benno Stein. 2020.
\newblock \href {https://doi.org/10.1145/3451964.3451978} {{The dilemma of the
  direct answer}}.
\newblock \emph{ACM SIGIR Forum}, 54(1):1--12.

\bibitem[{Purver et~al.(2001)Purver, Ginzburg, and Healey}]{purver:means}
Matthew Purver, Jonathan Ginzburg, and Patrick Healey. 2001.
\newblock On the means for clarification in dialogue.
\newblock In \emph{Proceedings of the 2nd SIGdial Workshop on Discourse and
  Dialogue}, Aalborg, Denmark.

\bibitem[{Rodr{\'i}guez and Schlangen(2004)}]{rodrischlang:catalog}
Kepa~Joseba Rodr{\'i}guez and David Schlangen. 2004.
\newblock \href
  {http://www.ling.uni-potsdam.de/~das/papers/rodrschl\_catalog.pdf} {Form,
  intonation and function of clarification requests in german task-oriented
  spoken dialogues}.
\newblock In \emph{Proceedings of Catalog (the 8th workshop on the semantics
  and pragmatics of dialogue; SemDial04)}, pages 101--108, Barcelona, Spain.

\bibitem[{Schegloff et~al.(1977)Schegloff, Jefferson, and
  Sacks}]{schegetal:prefself}
Emanuel~A. Schegloff, Gail Jefferson, and Harvey Sacks. 1977.
\newblock The preference for self-correction in the organisation of repair in
  conversation.
\newblock \emph{Language}, 53(2):361--382.

\bibitem[{Schlangen(2016)}]{schlangen:justification}
David Schlangen. 2016.
\newblock {Grounding, Justification, Adaptation: Towards Machines That Mean
  What They Say}.
\newblock In \emph{Proceedings of the 20th Workshop on the Semantics and
  Pragmatics of Dialogue (JerSem)}.

\bibitem[{Sellars(1969)}]{sellars:thought}
Wilfrid Sellars. 1969.
\newblock \href {https://doi.org/10.2307/2105537} {Language as thought and as
  communication}.
\newblock \emph{Philosophy and Phenomenological Research}, 29(4):506--527.

\bibitem[{Sellars(1954)}]{sellars:languagegames}
Wilfried Sellars. 1954.
\newblock Some reflections on language games.
\newblock \emph{Philosophy of Science}, 21:204--228.

\bibitem[{Sellars(1956)}]{sellars:epm}
Winfrid Sellars. 1956.
\newblock \emph{Empiricism and the Philosophy of Mind}.
\newblock Harvard University Press, Cambridge, Mass., USA.

\bibitem[{Shah and Bender(2022)}]{Shah2022}
Chirag Shah and Emily~M Bender. 2022.
\newblock {Situating Search}.
\newblock In \emph{CHIIR 2022}, pages 221--232.

\bibitem[{Takmaz et~al.(2022)Takmaz, Pezzelle, and
  Fern{\'a}ndez}]{takmaz-etal-2022-less}
Ece Takmaz, Sandro Pezzelle, and Raquel Fern{\'a}ndez. 2022.
\newblock \href {https://aclanthology.org/2022.cmcl-1.4} {Less descriptive yet
  discriminative: Quantifying the properties of multimodal referring utterances
  via {CLIP}}.
\newblock In \emph{Proceedings of the Workshop on Cognitive Modeling and
  Computational Linguistics}, pages 36--42, Dublin, Ireland. Association for
  Computational Linguistics.

\bibitem[{Vinyals et~al.(2015)Vinyals, Toshev, Bengio, and
  Erhan}]{vinyals:show}
Oriol Vinyals, Alexander Toshev, Samy Bengio, and Dumitru Erhan. 2015.
\newblock Show and tell: A neural image caption generator.
\newblock In \emph{Computer Vision and Pattern Recognition}.

\bibitem[{Whiting(2022)}]{whiting:crs}
Daniel Whiting. 2022.
\newblock \href {https://iep.utm.edu/} {Conceptual role semantics}.
\newblock In \emph{The Internet Encyclopedia of Philosophy}, ISSN 2161-0002.
  IEP.
\newblock Retrieved 2022-05-30.

\bibitem[{Wittgenstein(1984 {[1953]})}]{Witt:PU-corr}
Ludwig Wittgenstein. 1984 {[1953]}.
\newblock \emph{Tractatus Logicus Philosophicus und Philosophische
  Untersuchungen}, volume~1 of \emph{Werkausgabe}.
\newblock Suhrkamp, Frankfurt am Main.

\bibitem[{Zhou et~al.(2022)Zhou, Gopalakrishnan, Hedayatnia, Kim, Pujara, Ren,
  Liu, and Hakkani-Tur}]{zhou-etal-2022-think}
Pei Zhou, Karthik Gopalakrishnan, Behnam Hedayatnia, Seokhwan Kim, Jay Pujara,
  Xiang Ren, Yang Liu, and Dilek Hakkani-Tur. 2022.
\newblock \href {https://aclanthology.org/2022.acl-long.88} {Think before you
  speak: Explicitly generating implicit commonsense knowledge for response
  generation}.
\newblock In \emph{Proceedings of the 60th Annual Meeting of the Association
  for Computational Linguistics (Volume 1: Long Papers)}, pages 1237--1252,
  Dublin, Ireland. Association for Computational Linguistics.

\end{thebibliography}

\end{document}